\documentclass{article} 
\usepackage{iclr2025_conference,times}

\usepackage[utf8]{inputenc} 
\usepackage[T1]{fontenc}    

\usepackage{amsmath}        
\usepackage{amssymb}       
\usepackage{amsfonts}       
\usepackage{nicefrac}     

\usepackage{graphicx}     
\usepackage{color, colortbl, xcolor}
\usepackage{adjustbox}     
\usepackage{wrapfig}       
\usepackage{float}          
\usepackage{subfig}        

\usepackage{booktabs}      
\usepackage{multirow}      
\usepackage{makecell}       
\usepackage{array}         
\newcolumntype{C}[1]{>{\centering\arraybackslash}m{#1}} 

\usepackage{microtype}     
\usepackage{url}           
\usepackage{caption}      
\usepackage{framed}        
\usepackage{listings}      
\usepackage[shortlabels]{enumitem}

\definecolor{citecolor}{HTML}{2980b9}
\definecolor{linkcolor}{HTML}{c0392b}
\definecolor{darkorange}{HTML}{FF8C00}
\definecolor{chocolate}{HTML}{D2691E}
\definecolor{darkgreen}{HTML}{006400}
\definecolor{darkblue}{HTML}{00008B}
\definecolor{mediumblue}{HTML}{0000CD}
\definecolor{dodgerblue}{HTML}{1E90FF}
\definecolor{royalblue}{HTML}{4169E1}
\definecolor{shadecolor}{RGB}{237,237,237}
\definecolor{backred}{RGB}{255, 190, 190}
\definecolor{backblue}{RGB}{210, 230, 250}
\definecolor{zrrgreen}{HTML}{008000}
\definecolor{zrrblue}{HTML}{4682B4}
\definecolor{zrrred}{HTML}{B22222}
\definecolor{pink}{RGB}{240, 81, 121}
\definecolor{green}{RGB}{69, 189, 155}
\definecolor{yellow}{RGB}{253, 207, 110}
\definecolor{lightred}{RGB}{254, 129, 125}
\definecolor{lightblue}{RGB}{129, 184, 223}

\usepackage[hidelinks,breaklinks=true,colorlinks,bookmarks=false,
citecolor=citecolor,linkcolor=linkcolor]{hyperref}

\makeatletter
  \newcommand\figcaption{\def\@captype{figure}\caption}
  \newcommand\tabcaption{\def\@captype{table}\caption}
\makeatother

\title{MAVIS:\\Mathematical Visual Instruction Tuning with an Automatic Data Generation Engine}

\title{MAVIS:\\Mathematical Visual Instruction Tuning\\with an Automatic Data Engine}

\author{Renrui Zhang\thanks{Equal Contribution}\hspace{0.12cm}$^{1}$, Xinyu Wei$^{*2}$, Dongzhi Jiang$^{1}$, Ziyu Guo$^{1}$, Shicheng Li$^{2}$\vspace{0.1cm}\\ \textbf{Yichi Zhang$^{2}$, Chengzhuo Tong$^3$, Jiaming Liu$^2$, Aojun Zhou$^1$, Bin Wei$^{5}$}\vspace{0.1cm}\\ \textbf{Shanghang Zhang$^{2}$, Peng Gao$^{3}$, Chunyuan Li$^{4}$, Hongsheng Li$^{1}$}\vspace{0.3cm}\\
  $^1$CUHK\quad
  $^2$Peking University\quad
  $^3$Shanghai AI Laboratory\quad
  $^4$ByteDance\quad
  $^5$Oracle\vspace{0.1cm}\\
  \texttt{\{renruizhang, dzjiang, ziyuguo\}@link.cuhk.edu.hk}\\
}

\iclrfinalcopy 
\begin{document}

\maketitle

\begin{abstract}
Multi-modal Large Language Models (MLLMs) have recently showcased superior proficiency in general visual scenarios. However, we identify their mathematical capabilities remain under-explored with three areas to be improved: \textit{visual encoding of math diagrams}, \textit{diagram-language alignment}, and \textit{chain-of-thought (CoT) reasoning}. This draws forth an urgent demand for an effective training paradigm and a large-scale, comprehensive dataset with detailed CoT rationales, which is challenging to collect and costly to annotate manually. To tackle this issue, we propose \textbf{MAVIS}, a \textbf{MA}thematical \textbf{VIS}ual instruction tuning pipeline for MLLMs, featuring an automatic data engine to efficiently create mathematical visual datasets.
We design the data generation process to be entirely independent of human intervention or GPT API usage, while ensuring the diagram-caption correspondence, question-answer correctness, and CoT reasoning quality. With this approach, we curate two datasets, MAVIS-Caption (558K diagram-caption pairs) and MAVIS-Instruct (834K visual math problems with CoT rationales), and propose four progressive stages for training MLLMs from scratch.
First, we utilize MAVIS-Caption to fine-tune a math-specific vision encoder (CLIP-Math) through contrastive learning, tailored for improved diagram visual encoding. Second, we also leverage MAVIS-Caption to align the CLIP-Math with a large language model (LLM) by a projection layer, enhancing vision-language alignment in mathematical domains. Third, we adopt  MAVIS-Instruct to perform the instruction tuning for robust problem-solving skills, and term the resulting model as MAVIS-7B.
Fourth, we apply Direct Preference Optimization (DPO) to enhance the CoT capabilities of our model, further refining its step-wise reasoning performance.
On various mathematical benchmarks, our MAVIS-7B achieves leading results among open-source MLLMs, e.g., surpassing other 7B models by +9.3\% and the second-best LLaVA-NeXT (110B) by +6.9\%, demonstrating the effectiveness of our method. Code and data will be released \url{https://github.com/ZrrSkywalker/MAVIS}.
\end{abstract}

\section{Introduction}

The pursuit of artificial general intelligence necessitates models to seamlessly interpret and generate multi-modal data. In recent years, the advent of Large-language Models (LLMs)~\citep{brown2020language,touvron2023llama,touvron2023llama2,vicuna2023} and their Multi-modal extension (MLLMs)~\citep{zhang2024llamaadapter,gao2023llamaadapterv2,su2023pandagpt,ye2023mplugowl} have significantly facilitated this process across various fields, such as healthcare~\citep{singhal2023towards,shu2023visual}, autonomous driving~\citep{yang2023lidar,jin2024tod3cap}, and robotics~\citep{li2023manipllm, liu2024self}. Although MLLMs exhibit remarkable performance in diverse tasks and benchmarks, one arena where they have yet to fully demonstrate their potential is mathematical problem-solving in visual contexts.

\begin{figure*}[t!]
\centering
\includegraphics[width=\textwidth]{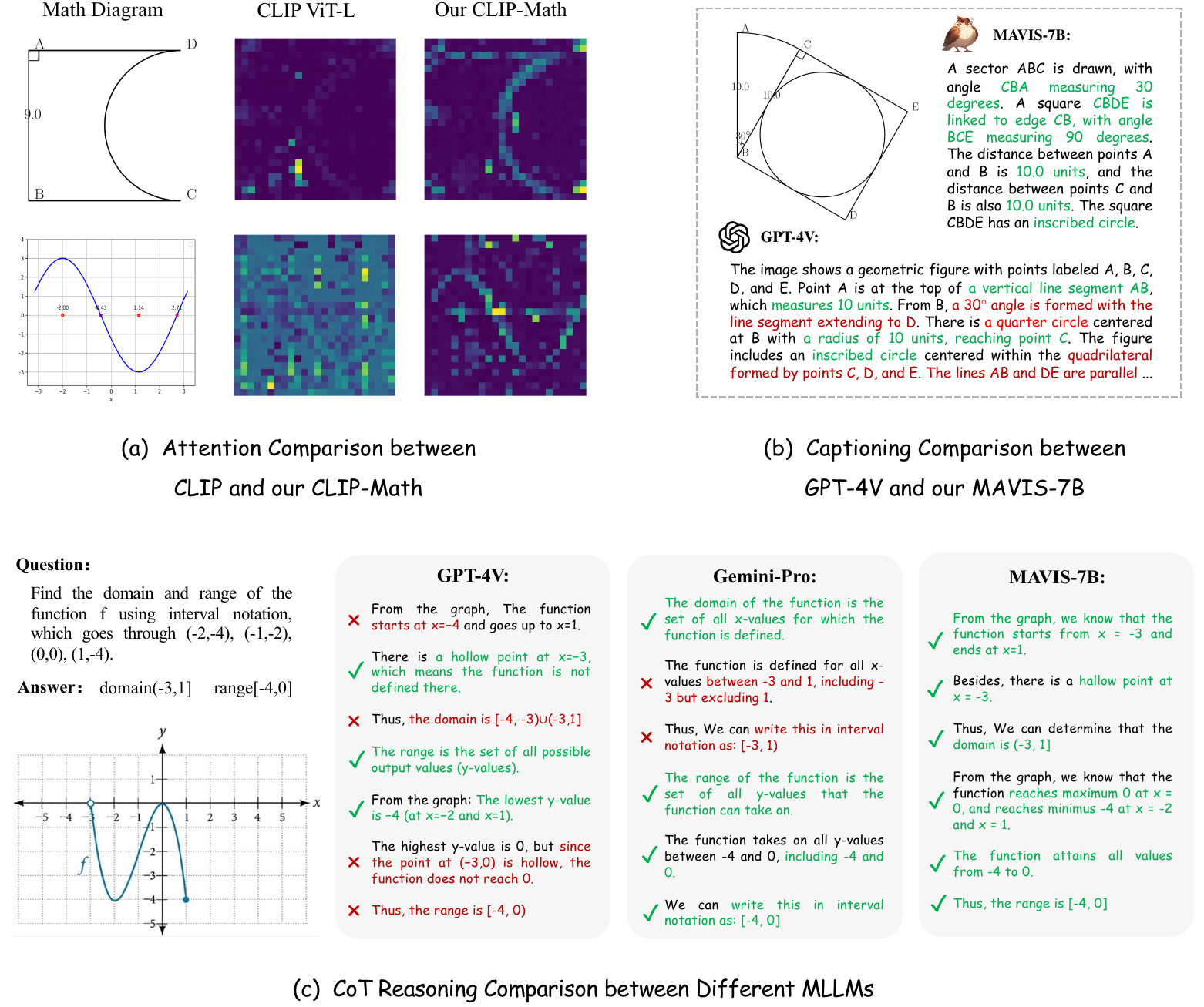}
   \caption{\textbf{(a)} We compare the attention map of class tokens from CLIP ViT-L~\citep{Radford2021LearningTV} and our CLIP-Math. Our vision encoder can better capture significant mathematical information within diagrams. \textbf{(b)} We compare the diagram captioning capabilities between GPT-4V~\citep{openai2023gpt4v} and our MAVIS-7B, where GPT-4V fall short of accurately recognizing mathematical elements. \textbf{(c)} We compare the chain-of-thought (CoT) reasoning between different models, showcasing that GPT-4V and Gemini-Pro~\citep{team2023gemini} suffer from low-quality reasoning process.}
\label{fig1}
\end{figure*}

Existing efforts~\citep{OpenAI2023GPT4TR,OpenAI2023ChatGPT,zhou2023solving} for text-only mathematics have attained considerable progress, largely attributed to the availability of sufficient and easily accessible training data. 
In contrast, solving visual mathematical problems remains a significant challenge for MLLMs, primarily due to the absence of a fully validated, effective training pipeline and the acute shortage of large-scale, high-quality datasets. Visual mathematical data is not only more costly to collect from publicly available sources compared to text-only data, but also requires expensive manual annotation to produce accurate step-by-step chain-of-thought (CoT) rationales integrating diagram information.
In light of these challenges, we identify three critical issues that impede the visual mathematical capabilities of MLLMs.

\begin{enumerate}[label=\roman*.]
    \item \textbf{Unsatisfactory math diagram embeddings by vision encoders.} Most MLLMs adopt a frozen CLIP~\citep{Radford2021LearningTV} as the vision encoder, which is pre-trained by natural images capturing real-world scenes with rich colors and textures. In contrast, math diagrams are composed of abstract curves, shapes, and symbols with a monochromatic color scheme, exhibiting large semantic gaps to general scenarios. As visualized in Figure~\ref{fig1} (a), the attention map of CLIP struggles to capture important information within math diagrams, which cannot provide satisfactory visual embeddings for LLMs to understand. 
    
    \item \textbf{Diagram-language misalignment between vision encoders and LLMs.} Likewise, the vision-language pre-training stage of MLLMs also adopts natural image-caption pairs for cross-modal alignment. Due to the domain gap, while they can generate accurate captions for real-world images, but fall short of recognizing basic mathematical elements and narrating their relations. As compared in Figure~\ref{fig1} (b), even GPT-4V~\citep{openai2023gpt4v} produces low-quality descriptions for simple geometric figures and functions, indicating LLMs are not well aligned with the visual embedding space of math diagrams.
    
    \item \textbf{Inaccurate CoT reasoning capabilities with visual elements by MLLMs.} Referring to the CoT evaluation in MathVerse~\citep{zhang2024mathverse}, incorporating the diagram input would adversely affect the reasoning quality of MLLMs compared to using only the text-only question. As visualized in Figure~\ref{fig1} (c), we observe the problem-solving process of GPT-4V and Gemini-Pro~\citep{team2023gemini} both suffer from low-quality CoT reasoning accuracy. This demonstrates the incapability of MLLMs to leverage visual cues for precise step-by-step mathematical problem-solving. 
    
\end{enumerate}

Therefore, to mitigate these issues, it is essential to develop an extensive dataset and effective training approach tailored to visual mathematics. In this paper, we propose \textbf{MAVIS}, a \textbf{MA}thematical \textbf{VIS}ual instruction tuning paradigm and an automatic data generation engine for MLLMs, which aims to fully unleash their potential for diagram visual encoding and reasoning capabilities. We introduce two meticulously curated datasets, a progressive four-stage training pipeline, and a visual mathematical specialist, MAVIS-7B. We summarize the contributions of our work as follows.

\begin{itemize}
    \item \textbf{Automatic Mathematical Visual Data Engine.} 
    To eliminate the need for labor-intensive annotation and expensive GPT API~\citep{openai2023gpt4v,OpenAI2023GPT4TR} usage, we designed our data engine to be entirely rule-based and fully automated. This engine handles every aspect of mathematical data creation, including diagram drawing, caption generation, question-answer synthesis, and CoT rationale production. With this approach, we curate two large-scale, high-quality mathematical visual datasets, MAVIS-Caption and MAVIS-Instruct, widely covering plane geometry, analytic geometry, and function.
    MAVIS-Caption consists of 558K diagram-caption pairs automatically created by our data engine with accurate vision-language correspondence.
    MAVIS-Instruct includes 834K visual math problems, which includes 582K data constructed by our data engine and additional 252K data augmented by GPT-4V from manual collection and existing datasets~\citep{chen2021geoqa,lu2021inter}. Each problem is annotated with a CoT rationale, and modified to contain minimized textual redundancy that enforces MLLMs to pay more attention on visual diagrams.
    
    \item \textbf{Four-stage Training Pipeline.} Our training framework involves four progressive stages designed to sequentially address the aforementioned identified deficiencies in MLLMs. Firstly, we utilize MAVIS-Caption to fine-tune a math-specific vision encoder by contrastive learning, termed CLIP-Math, to enable better visual representations of math diagrams. Subsequently, we align this encoder with the LLM to ensure effective diagram-language integration also by MAVIS-Caption. After that, our MAVIS-Instruct is adopted to instruction-tune the MLLM, which provides sufficient step-wise problem-solving supervision. Finally, we employ Direct Preference Optimization (DPO)~\citep{dpo_rafailov2024directpreferenceoptimizationlanguage} with annotated CoT rationales in MAVIS-Instruct to further enhance the reasoning capabilities of our model.
    
    \item \textbf{Mathematical Visual Specialist.}
    After the four-stage training, we develop MAVIS-7B, an MLLM specifically optimized for visual mathematical problem-solving. On various evaluation benchmarks, our model achieves leading performance compared to existing open-source MLLMs, e.g., surpassing other 7B models by +9.3\% and the second-best LLaVA-NeXT (110B)~\citep{li2024llavanext-strong} by +6.9\% on MathVerse~\citep{zhang2024mathverse}. The quantitative results and qualitative analysis both validate the significance of our approach.


\end{itemize}

\section{Automatic Data Engine}

To cope with the substantial data requirements of MLLMs, it is essential to have access to extensive training instances. However, for visual mathematics, the paucity of publicly available datasets poses a challenge, and creating such data manually is also not feasible due to the high cost involved. Therefore, we develop an automatic data engine to efficiently generate high-quality math diagrams (Section~\ref{s3.1}), detailed captions (Section~\ref{s3.2}), and question-answer pairs with CoT rationales (Section~\ref{s3.3}).

\subsection{Diagram Generation}
\label{s3.1}

Covering most mathematical scenarios, we adopt three diagram types: plane geometry, analytic geometry, and function. Note that all the logic of the data engine is implemented in Python, and we employ Matplotlib for the graphical rendering of the diagrams.

\paragraph{Plane Geometry Diagram.}
As such diagrams typically consist of spatial combinations of various basic shapes, we utilize principles from multi-hop data curation to develop customized generation rules. These rules allow for the iterative integration of new shapes into existing configurations. Initially, we establish a core set of shapes, including squares, rectangles, triangles, sectors, etc, for diagram generation. Starting with a randomly selected shape, we extend another shape from the set along one of its straight sides. By iterating this process, we can construct diverse plane geometry diagrams featuring different combinations of shapes. Additionally, we randomly label the vertices with letters (e.g., A, B, C) and annotate numerical values relevant to geometric properties (e.g., side lengths and angles), simulating realistic plane geometry problems.

\paragraph{Analytic Geometry Diagram.}
Likewise, our approach begins by defining a basic figure set that differs slightly from that used in plane geometry; for example, we include additional elements such as points and line segments. We then construct a Cartesian coordinate system, complete with grid lines and scaled axes. The range of the coordinate system is randomly determined within a predefined scope. Subsequently, we select a number from 1 to 3 to indicate the number of figures to be drawn on the graph, and randomly choose coordinates for the top-left vertices to plot these figures at varied sizes (using these points as centers for circles). Unlike plane geometry, we ensure that the figures do not overlap, except for points and segments, and maintain the figure areas within a suitable scale.

\paragraph{Function Diagram.}
We focus on seven fundamental function types: polynomial, sine, cosine, tangent, logarithmic, absolute value, and piece-wise polynomial functions. For each function type, we parameterize the equations with random variables, such as coefficients and constants within a predefined range (e.g., $a$ and $b$ in $y=ax+b$), which facilitates the generation of diverse function graphs. We also adopt the same Cartesian coordinate system employed for analytic geometry. Additionally, for specific caption or question-answering samples, we also plot key features like extreme points and zero points of the functions, providing additional visual information that aids in the understanding and reasoning of these mathematical functions.

\begin{figure*}[t]
\centering
\begin{minipage}[c]{0.43\textwidth}
\small
\centering
\tabcaption{\textbf{Statistics of MAVIS-Caption.}}
\vspace{0.1cm}
\label{t1}
\centering
\begin{adjustbox}{width=\linewidth}
\begin{tabular}{lr}
 \toprule
 \textbf{Statistic} & \textbf{Number} \\
 \midrule
  \textit{Total Captions}&  \\
  ~- Total number &588K \\
  ~- Average length (words) &62.85  \\
  ~- Average length (characters) &339.68  \\
  ~- Vocabulary size &418  \\
 \midrule
 \textit{Plane Geometry} &  \\
  ~- Total number &299K (50.9\%) \\
  ~- Average length (words) &69.77  \\
  ~- Average length (characters) &385.85  \\
  ~- Vocabulary size &195  \\
  \midrule
  \textit{Analytic Geometry} &  \\
  ~- Total number & 77K (13.1\%) \\
  ~- Average length (words) & 39.64 \\
  ~- Average length (characters) & 210.10 \\
  ~- Vocabulary size & 158 \\
 \midrule
  \textit{Function} &  \\
  ~- Total number & 212K (36.0\%) \\
  ~- Average length (words) & 61.48 \\
  ~- Average length (characters) & 321.46 \\
  ~- Vocabulary size & 149 \\
 \bottomrule
 \end{tabular}
 \end{adjustbox}
\end{minipage}
\quad
\begin{minipage}[c]{0.524\textwidth}
\small
\centering
\tabcaption{\textbf{Subject Distribution of MAVIS-Instruct.}}
\vspace{0.1cm}
\label{t2}
\begin{adjustbox}{width=\linewidth}
\begin{tabular}{lr}
 \toprule
 \textbf{Statistic} & \textbf{Number} \\
 \midrule
  \textit{Total questions} & 834K \\ 
  ~- Multiple-choice questions & 615K (62.4\%) \\ 
  ~- Free-form questions & 218K (37.6\%) \\ 
 \midrule
 \textit{Data Engine Generated Problems} & 582K \\
  ~- Geometry questions & 466K (80.0\%) \\
  ~- Function questions & 116K (20.0\%) \\
\midrule
  \textit{Data Engine Captions Annotated by GPT-4} & 51K \\
  ~- Geometry questions & 30K (58.8\%) \\
  ~- Function questions & 21K (41.2\%) \\
\midrule
  \textit{Manual Collection Augmented by GPT-4}& 83K \\
  ~- Geometry questions & 72K (86.5\%) \\
  ~- Function questions & 11K (13.5\%) \\
 \midrule
 \textit{Existing Datasets Augmented by GPT-4} & 118K \\
  ~- Geometry questions & 118K (100.0\%)\\
  ~- Function questions & 0 (0\%) \\
 \midrule
 Number of unique images & 611K (73.3\%) \\
 Number of unique questions & 804K (96.5\%) \\
 Number of unique answers & 675K (81.0\%) \\
 \midrule
 Average question length & 44.60\\
 Average answer length & 62.82 \\
 \bottomrule
 \end{tabular}
 \end{adjustbox}
\end{minipage}
\end{figure*}


\subsection{MAVIS-Caption}
\label{s3.2}

\begin{figure*}[t!]
\centering
\includegraphics[width=0.98\textwidth]{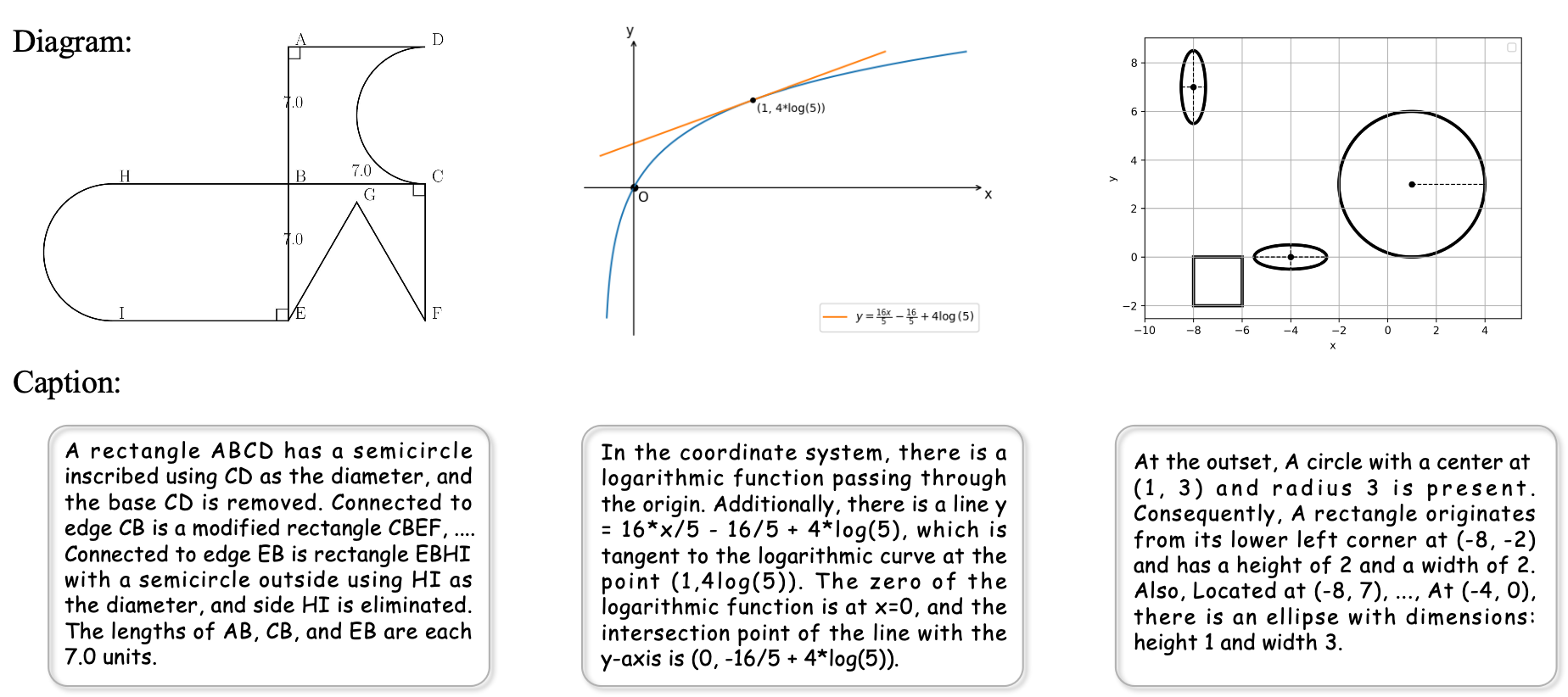}
   \caption{\textbf{MAVIS-Caption Dataset.} We showcase three diagram-caption pairs of plane geometry, function, and analytic geometry in MAVIS-Caption, generated by our developed data engine.}
\label{fig2}
\end{figure*}

With our mathematical visual data engine, we first curate a diagram-caption dataset, MAVIS-Caption, as shown in Figure~\ref{fig2}, aiming to benefit the diagram visual representations and cross-modal alignment.

\paragraph{Data Overview.}
As presented in Table~\ref{t1}, the MAVIS-Caption dataset comprises 588K diagram-caption pairs. This includes 299K for plane geometry, 77K for analytic geometry, and 212K for function. The average word length of the captions is 61.48 words, reflecting their detailed descriptive nature. The overall vocabulary size is 149, indicating the diversity in language expression. We adopt different strategies to generate captions for three types of diagrams. It is important to note that GPT-4~\citep{OpenAI2023GPT4TR} is only utilized during the template creation stage; it is not used at any point during the automatic caption generation process.

\paragraph{Plane Geometry Caption.}
We follow the iterative geometric generation process to develop regulations for an accurate and detailed caption. We first prompt GPT-4 to create three sets of language templates: the descriptive content for fundamental shapes (e.g., \textit{``A Triangle \{\} with two congruent sides \{\} and \{\}''}), the phrases to denote specific attributes (e.g., \textit{``Angle \{\} measures \{\} degrees''}), and the conjunction to link two adjacent shapes (e.g., \textit{``Attached to edge \{\} of shape \{\}, there is a \{\}''}). Then, based on various generation scenarios, we fill and merge these templates to acquire a coherent description of the geometric figure.
    
\paragraph{Function Caption.}
As function diagrams typically showcase a single curve, we directly utilize GPT-4 to generate templates describing various properties of functions, including expressions, domains, ranges, extreme points, and zero points. Each template is then filled based on specific cases, such as \textit{``The expression of the function is $y = -3x^3 - 2x^2 - 2x - 2$. Within the range of x values $[-3.0, 4.0]$, zero points occur at $-0.83$ ...''}.

\paragraph{Analytic Geometry Caption.} We also employ GPT-4 to obtain two sets of language templates: the description of coordinates and attribute information for basic figures (e.g., \textit{``The square with its base left corner at \{\} features sides of \{\} in length''}) and the spatial relation for nearby figures (e.g., \textit{``On the bottom right of \{\}, there is a \{\}''}). The captions are then formulated by filling in the coordinates and selecting appropriate spatial relationship templates through coordinate comparison.
    

\subsection{MAVIS-Instruct}
\label{s3.3}

\begin{figure*}[t!]
\centering
\includegraphics[width=0.999\textwidth]{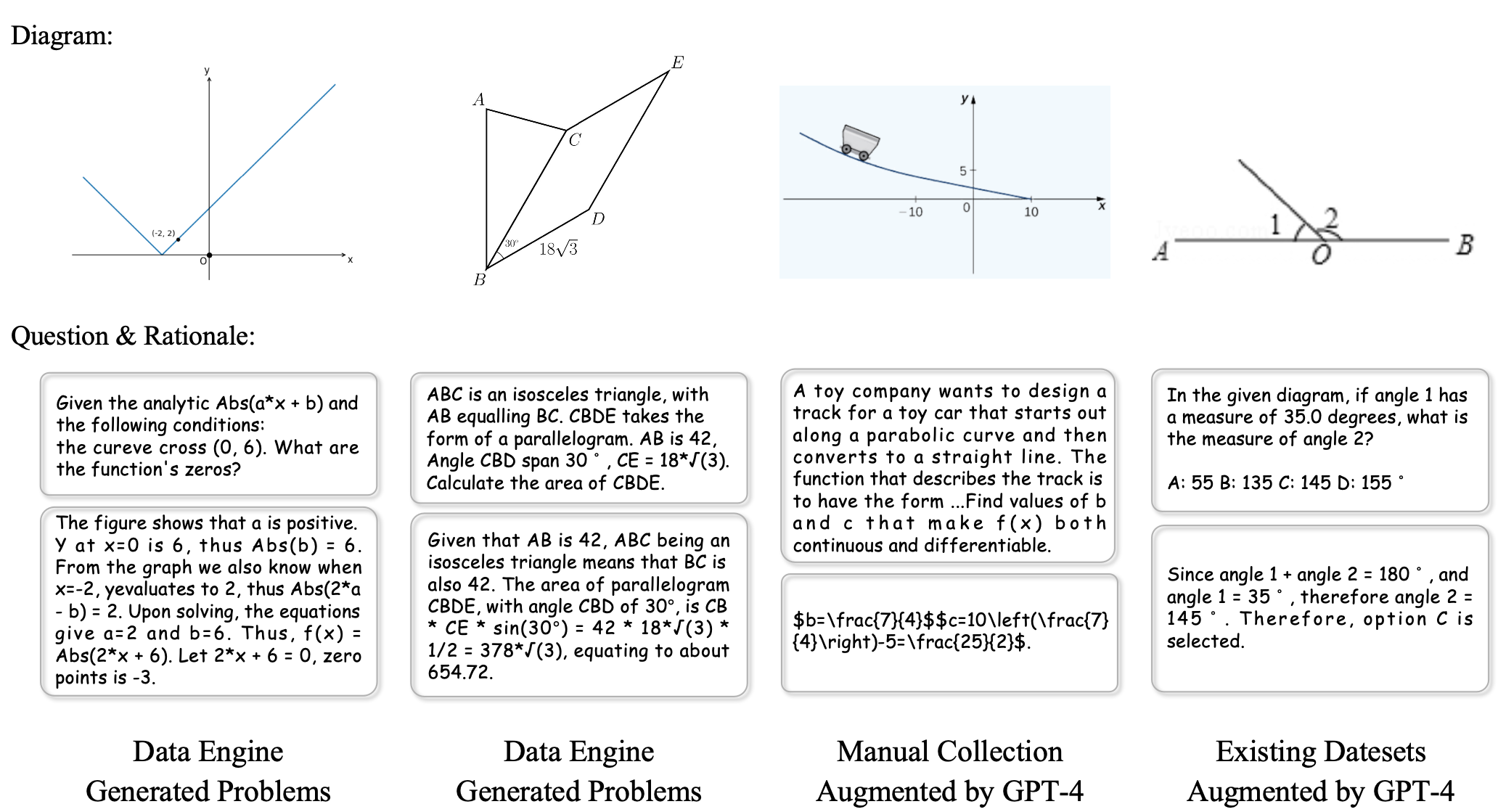}
   \caption{\textbf{MAVIS-Instruct Dataset.} We showcase the generated visual math problems from four sources within MAVIS-Instruct, which contain detailed rationales and minimized textual redundancy.}
\label{fig3}
\end{figure*}

Besides the diagram-caption data, we curate MAVIS-Instruct of extensive problem-solving data, which endows MLLMs with visual mathematical reasoning capabilities and serve as the basis for Direct Preference Optimization (DPO)~\citep{dpo_rafailov2024directpreferenceoptimizationlanguage}, as shown in Figure~\ref{fig3}.

\paragraph{Data Overview.}
As illustrated in Table~\ref{t2}, the MAVIS-Instruct dataset consists of a total of 834K visual math problems. Given that the proportion of analytic geometry problems is relatively small, we classify them with function problems for simplicity. Each problem in MAVIS-Instruct includes a CoT rationale providing step-by-step solutions, with an average answer length of 150 words. We have minimized textual redundancy in the questions, eliminating unnecessary contextual information, distracting conditions, and attributes readily observable from the diagrams. This reduction in text forces MLLMs to enhance their capability to extract essential content from visual inputs. MAVIS-Instruct is assembled from four distinct sources to ensure broad coverage.

\paragraph{Data Engine Generated Problems.} Within our data engine, we manually craft rigorous regulations to produce visual math problems with accurate CoT annotations. Similar to caption generation, GPT API is not involved in the automatic synthesis process of questions, answers, and CoT rationales.
\begin{itemize}
    \item \textbf{Plane Geometry Problems.} We initially prompt GPT-4 to compile a comprehensive set of mathematical formulas applicable to each basic shape (e.g., Pythagorean theorem for right triangles and area formula for circles). Then, for a geometric diagram, we randomly select a known condition within a shape as the final solution target, and systematically deduce backward to another condition, either within the same shape or an adjacent one, using a randomly selected mathematical formula. This deduced condition is then set as unknown, and we continue iterative backward deductions as necessary. The final condition, along with any conditions in the last step, are presented as initial attributes in the question. The rationales can be simply obtained by reversing this backward deduction process.
    \item \textbf{Function Problems.} As the properties of functions are predetermined, we utilize GPT-4 to generate diverse reasoning templates. These templates facilitate the solving of one function property based on other provided properties, thereby ensuring the generation of high-quality function rationales. The related function properties include analytical expression, function values, zeros, extremum points, monotonicity, derivatives, and integrals. 
    To accurately reason these properties, the CoT annotation incorporates understanding of function types, solving the analytical expressions of equations, and interpreting function graphs.

\end{itemize}

\paragraph{Data Engine Captions Annotated by GPT-4.} Given the detailed captions and diagrams generated by our data engine, we can prompt GPT-4V with these sufficient conditions to synthesis question-answering data and ensure its correctness. We first generate a new set of 17K diagram-caption pairs that do not overlap with the previous MAVIS-Caption, which avoids answer leakage within the detailed caption. Then, we prompt GPT-4V to generate 3 new problems with rationales, obtaining 51K data in total from the diagram-caption pairs.

\paragraph{Manual Collection Augmented by GPT-4.} To incorporate high-quality problems found in real-world contexts, we manually collect 4K math problems with diagrams from publicly available resources. Recognizing that these sources often lack detailed rationales and may contain redundant text, we initially utilize GPT-4V to annotate a detailed solving process and streamline the question text to reduce redundancy. Subsequently, for each collected instance, we input the question, rationale, and diagram into GPT-4 and employ customized few-shot prompts to generate 20 new problems per original, comprising 15 multiple-choice questions and 5 free-form questions. This process contributes a total of 83K problems to the dataset.

\paragraph{Existing Datasets Augmented by GPT-4.} Given existing well-organized geometric datasets, we can also leverage them to expand MAVIS-Instruct. Referring to previous prompt designs, we augment the 8K training set from two dataset, Geometry-3K~\citep{lu2021inter} and GeoQA+~\citep{Chen2021GeoQAAG}, into 80K visual problems with accompanying rationales, mapping each original problem to 10 new ones. Due to the scarcity of publicly available function data, we do not include function problems from this source.

\section{Mathematical Visual Training}

\begin{figure*}[t!]
\centering
\includegraphics[width=\textwidth]{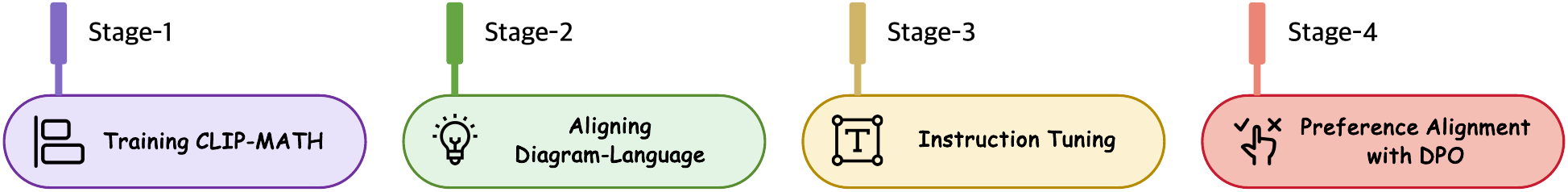}
   \caption{\textbf{Four-stage Training Pipeline of MAVIS.} With our curated MAVIS-Caption and MAVIS-Instruct, we adopt four progressive stages for training a mathematical visual specialist from scratch.}
\label{fig3.3}
\end{figure*}

With the curated datasets, we devise a four-stage training pipeline for endowing MLLMs with mathematical visual capabilities as shown in Figure~\ref{fig3.3}. They respectively aim to mitigate the three deficiencies within existing MLLMs, i.e., diagram visual encoding, diagram-language alignment, and mathematical reasoning skills in visual contexts.

\subsection{Stage 1: Training CLIP-Math}

To enhance CLIP's~\citep{Radford2021LearningTV} inadequate visual encoding of math diagrams, we utilize MAVIS-Caption to train a specialized CLIP-Math encoder. Specifically, we fine-tune a pre-trained CLIP-Base model following the conservative learning scheme. The math diagrams are fed into the learnable vision encoder, while the corresponding captions are processed by the text encoder, which remains frozen to provide reliable supervision. Via contrastive training, the model learns to adapt from its original natural image domain to mathematical contexts, increasing its focus on essential visual elements within diagrams, as demonstrated in Figure~\ref{fig1} (a). The optimized CLIP-Math encoder now delivers more precise and robust representations of math diagrams, establishing a solid foundation for the subsequent visual interpretation of LLMs.

\subsection{Stage 2: Aligning Diagram-language}

After acquiring the CLIP-Math encoder, we further integrate it with LLMs using MAVIS-Caption to boost cross-modal alignment between math diagrams and language embedding space. Using a simple two-layer MLP as the projection layer, we transform the visual encodings from CLIP-Math, and prepend them as a prefix to the LLM input. 
This process, guided by the diagram captioning task, enables the LLM to accurately recognize mathematical components and spatial arrangements. With the diagram-language alignment, LLMs are equipped with the interpretation capability in math diagrams, serving as an initial step toward deeper mathematical reasoning.
In this stage, we freeze the CLIP-Math, and train the projection layer along with the LoRA-based~\citep{hu2021lora} LLM.


\begin{table*}[!t]
\small
\centering
\caption{\textbf{Evaluation on MathVerse's \textit{testmini} Set with Six Problem Versions.} `CoT-E' and `Acc' denote the scores of CoT evaluation strategy and the scores of direct `true or false' accuracy, respectively. `$^*$' denotes previous mathematical visual specialists. The highest scores for \colorbox{backred!50}{closed-source} and \colorbox{backblue!75}{open-source} MLLMs are marked in red and blue respectively.}
\begin{adjustbox}{width=\linewidth}
    \begin{tabular}{l|C{0.8cm}|C{0.9cm}C{0.9cm}|C{0.9cm}C{0.9cm}|C{0.9cm}C{0.9cm}|C{0.9cm}C{0.9cm}|C{0.9cm}C{0.9cm}|C{0.9cm}C{0.9cm}}
    \toprule
    \multirow{3}*{\makecell*[l]{\large Model}} 
    &\multirow{3}*{\makecell*[c]{LLM\\Size}}    
    &\multicolumn{2}{c|}{\makecell*[c]{All}}
    &\multicolumn{2}{c|}{\makecell*[c]{\shortstack{\vspace*{0.1pt}\\Text\\\vspace*{0.2pt}\\Dominant}}} 
    &\multicolumn{2}{c|}{\makecell*[c]{\shortstack{\vspace*{0.1pt}\\Text\\\vspace*{0.2pt}\\Lite}}}
    &\multicolumn{2}{c|}{\makecell*[c]{\shortstack{\vspace*{0.1pt}\\\ Vision\ \ \\\vspace*{0.2pt}\\Intensive}}}
    &\multicolumn{2}{c|}{\makecell*[c]{\shortstack{\vspace*{0.1pt}\\\ Vision\ \ \\\vspace*{0.2pt}\\Dominant}}}
    &\multicolumn{2}{c}{\makecell*[c]{\shortstack{\vspace*{0.1pt}\\\ Vision\ \ \\\vspace*{0.2pt}\\Only}}}\\
    \cmidrule{3-14}
    &  & CoT-E & \ Acc\   & CoT-E & Acc& CoT-E & Acc& CoT-E & Acc& CoT-E & Acc& CoT-E & Acc  \\
    \midrule
    \multicolumn{14}{c}{\textit{Baselines}}\\
    \cmidrule{1-14}
    Random Chance & - & - & 12.4 & - & 12.4  & - & 12.4 & - & 12.4 & - & 12.4 & - & 12.4    \\
    Human         & - & - & 64.9 & - & 71.2  & - & 70.9 & - & 61.4 & - & 68.3 & - & 66.7     \\
    \cmidrule{1-14}
    \multicolumn{14}{c}{\textit{LLMs}}\\
    \cmidrule{1-14}
    ChatGPT &- &- &-  & 51.3 & 33.3  & 38.5 & 18.9 &- &-  &-& - & - &- \\
    GPT-4   &- &- &-  & 63.4 & 46.5  & 40.7 & 20.7 &- &-  &-& - & - &- \\
    \cmidrule{1-14}
    \multicolumn{14}{c}{\textit{Closed-source MLLMs}}\\
    \cmidrule{1-14}
    Qwen-VL-Plus & - & 21.3 & 11.8 &26.0&15.7&21.2&11.1&18.5&9.0& 19.1 & 13.0&21.8& 10.0\\
    Gemini-Pro   & - & 35.3 & 23.5 & 39.8 & 26.3  & 34.7 & 23.5 & 32.0 & 23.0 & 36.8 & 22.3 & 33.3 & 22.2 \\
    Qwen-VL-Max  & - & 37.2 & 25.3 & 42.8 & 30.7  & 37.7 & 26.1 & 33.6 & 24.1 & 35.9 & 24.1 & 35.9 & 21.4 \\
    GPT-4V       & - &\colorbox{backred!50}{54.4} &\colorbox{backred!50}{39.4} &\colorbox{backred!50}{63.1} &\colorbox{backred!50}{54.7} &\colorbox{backred!50}{56.6} &\colorbox{backred!50}{41.4} &\colorbox{backred!50}{51.4} &\colorbox{backred!50}{34.9} &\colorbox{backred!50}{50.8} &\colorbox{backred!50}{34.4} &\colorbox{backred!50}{50.3} &\colorbox{backred!50}{31.6}\\
    \cmidrule{1-14}
    \multicolumn{14}{c}{\textit{Open-source MLLMs}}\\
    \cmidrule{1-14}
    LLaMA-Adapter-V2 & 7B & 5.8 & 5.7 &7.8&6.2&6.3&5.9&6.2&6.1& 4.5 & 4.2&4.4 & 6.1\\
    ImageBind-LLM    & 7B & 10.0 & 9.2 & 13.2&11.4&11.6&11.3&9.8&8.9& 11.8 & 11.2&3.5& 3.4 \\
    mPLUG-Owl2       & 7B & 10.3 & 5.9 & 11.6&6.6&11.4&6.3&11.1&6.3& 9.4 & 5.6&8.0 & 4.9\\
    LLaVA-1.5        & 7B & 12.7 & 7.6  & 17.1 & 8.8 & 12.0 & 7.6 & 12.6 & 7.4 & 12.7 & 7.4 & 9.0 & 6.9\\
    SPHINX-Plus      & 13B & 14.0 &12.2& 16.3 & 13.9 &12.8&11.6&12.9&11.6& 14.7 & 13.5 & 13.2 & 10.4 \\
    G-LLaVA$^*$          & 7B & 15.7 & 16.6 &22.2&20.9&20.4&{20.7}&16.5&{17.2}& 12.7 & 14.6&6.6 & 9.4 \\
    LLaVA-NeXT       & 8B & 17.2 & 15.6 &21.6&19.4&19.7&15.2&17.6&16.8& 14.9 & 15.2 &12.1 & 11.3 \\
    ShareGPT4V       & 13B & 17.4 & 13.1 &21.8&16.2&20.6&16.2&18.6&15.5&16.2&13.8& 9.7 & 3.7\\
    SPHINX-MoE       & 8$\times$7B & 22.8 & 15.0 & 33.3 & 22.2 & 21.9 & 16.4 & 21.1 & 14.8 & 19.6 & 12.6 & 18.3 & 9.1\\
    Math-LLaVA$^*$       & 13B & 24.1 &  19.0 & 34.2 & 21.2 & 22.7 & 19.8 & 21.1 & 20.2 & 20.3 & 17.6 & 22.2 & 16.4\\
    InternLM-XC2.    & 7B & 25.9 & 16.5 & 36.9 & 22.3 & 28.3 & 17.0 & 20.1 & 15.7 & 24.4 & 16.4 & 19.8 & 11.0\\
    LLaVA-NeXT       & 110B & 28.3 & 24.5 & 37.1 & 31.7 & 29.1 & 24.1 & 22.6 & 21.0 & 21.8 & 22.1 & 30.9 & \colorbox{backblue!75}{20.7}\\
    \cmidrule{1-14}
    \makecell[l]{\textbf{MAVIS-7B w/o DPO}}       & 7B & 33.7 & 27.5 & 42.5 & 41.4 & 36.3 & 29.1 & 33.3&  27.4 & 29.3 & 24.9 & 27.1 & 14.6 \\
    \makecell[l]{\textbf{MAVIS-7B}}         & 7B & \colorbox{backblue!75}{35.2} & \colorbox{backblue!75}{28.4} & \colorbox{backblue!75}{43.2} & \colorbox{backblue!75}{41.6} & \colorbox{backblue!75}{37.2} & \colorbox{backblue!75}{29.5} & \colorbox{backblue!75}{34.1}&  \colorbox{backblue!75}{27.9} & \colorbox{backblue!75}{29.7} & \colorbox{backblue!75}{24.7} & \colorbox{backblue!75}{31.8} & 18.3 \\
    \bottomrule
    \end{tabular}
\end{adjustbox}
\label{t10}
\end{table*}

\subsection{Stage 3: Instruction Tuning}
On top of that, we leverage MAVIS-Instruct to endow MLLMs with CoT reasoning and problem-solving capabilities in visual mathematics. The detailed rationales within each problem's solution provide high-quality reasoning guidance for MLLMs, significantly enhancing their step-by-step CoT process. Furthermore, as we have minimized the redundancy within question texts during the construction process, such text-lite problem formats, referring to MathVerse~\citep{zhang2024mathverse}, facilitate MLLMs to capture more essential information from the visual embeddings for problem-solving, rather than relying on shortcuts to only process the textual content. In this stage, we unfreeze both the projection layer and apply LoRA~\citep{hu2021lora} for the LLM for a thorough tuning.

\subsection{Stage 4: Preference Alignment with DPO}
After the instruction tuning phase, the resulting model gains the capability for CoT reasoning on visual math problems. However, it may still produce inaccurate intermediate steps due to insufficient supervision for generating the best reasoning path. To address this, we further apply CoT preference alignment using the DPO~\citep{dpo_rafailov2024directpreferenceoptimizationlanguage} algorithm to further enhance the model's reasoning performance. Specifically, we adopt the instruction-tuned model to first infer CoT reasoning process on the 582K problems generated by data engine within MAVIS-Instruct. Then, we filter out the incorrect outputs (88K data) based on the final answer as the negative reasoning samples in DPO, and directly utilize the annotated CoT process as the positive samples. We only unfreeze the LoRA parameters for DPO training, and finally obtain our mathematical specialist, MAVIS-7B.

\section{Experiment}

We first detail our experimental settings in Section~\ref{s5.1}, and then discuss the quantitative on different benchmarks and qualitative examples in Sections~\ref{s5.2} and~\ref{s5.3}, respectively. Please refer to the Appendix for more data details and ablation studies.

\subsection{Experimental Settings}
\label{s5.1}

\paragraph{Implementation Details.} We adopt a CLIP ViT-L~\citep{Radford2021LearningTV} as the pre-trained model to fine-tune our CLIP-Math, and utilize Mammoth2-7B~\citep{yue2024mammoth2} as the base LLM to construct MAVIS-7B. In the first stage, we fine-tune the CLIP for 10 epochs with a batch size 16 and an initial learning rate $2e^{-6}$. In the second stage, we train the diagram-language alignment for 1 epoch with a batch size 32 and an initial learning rate $2e^{-6}$, and adopt LoRA~\citep{hu2021lora} with a rank 128. In the third and fourth stages, we adopt the same training settings as the second one.

\paragraph{Evaluation Schemes.}
We evaluate our model MAVIS-7B on several popular mathematical benchmarks, MathVerse~\citep{zhang2024mathverse}, GeoQA~\citep{chen2021geoqa}, FunctionQA (function problems in MathVista~\citep{Lu2023MathVistaEM}), MMMU-Math (the math problems in MMMU~\citep{yue2023mmmu}), MathVision~\citep{wang2024measuring}, three mathematical categories in MathVista, and We-Math~\citep{qiao2024we}. We compare a variety of existing MLLMs, including two mathematical visual specialist~\citep{gao2023g,shi2024math}, two LLMs~\citep{OpenAI2023ChatGPT,OpenAI2023GPT4TR}, and other general MLLMs~\citep{bai2023qwen,gao2023llamaadapterv2, ye2023mplugowl2,liu2023improvedllava,Chen2023ShareGPT4VIL,gao2024sphinx,dong2024internlm,liu2024llavanext,chen2023minigpt,gao2024sphinx}.

\begin{table*}[t]
    \centering
    \small
    \caption{\textbf{Evaluation on Six Mathematical Benchmarks.} `MMMU-Math' denotes the math problems within the test set of MMMU. `GPS', `ALG', and `GEO' denote geometry problem solving, algebraic, and geometry in MathVista's \textit{testmini} set. `S1', `S2', and `S3' denote different problem steps in We-Math's \textit{testmini} set. `$^*$' denotes previous mathematical visual specialists. The highest scores for \colorbox{backred!50}{closed-source} and \colorbox{backblue!75}{open-source} MLLMs are marked in red and blue respectively.}
    \label{tab:mavis_comparison}
    \begin{adjustbox}{width=\linewidth}
    \begin{tabular}{l|c|cccccccccc}
        \toprule
         \multirow{2}*{\makecell*[l]{Model}} &\multirow{2}*{\makecell{LLM\\Size}} & \multirow{2}*{GeoQA} & \multirow{2}*{FunctionQA} & \multirow{2}*{MMMU-Math} & \multirow{2}*{MathVision} & \multicolumn{3}{c}{MathVista} &\multicolumn{3}{c}{We-Math} \\
         &&&&&&\tiny GPS&\tiny ALG&\tiny GEO&\tiny S1&\tiny S2&\tiny S3\\
         \cmidrule{1-12}
         \multicolumn{12}{c}{\textit{Baselines}}\\
         \cmidrule{1-12}
         Random Chance &- & 17.1 & - & 21.6 & 7.2  & 24.1 & 25.8 & 22.7 & - & - & -\\
         Human         &- & 92.3 & - & 84.2 & 68.8 & 48.4 & 50.9 & 51.4 & - & - & -\\
         \cmidrule{1-12}
         \multicolumn{12}{c}{\textit{LLMs}}\\
         \cmidrule{1-12}
         ChatGPT &- & - & - & -    & 9.7  & 31.7 & 32.4 & 33.0 & - & - & -\\
         GPT-4   &- & - & - & 30.6 & 13.1 & 31.7 & 33.5 & 32.2 & - & - & -\\
         \cmidrule{1-12}
         \multicolumn{12}{c}{\textit{Closed-source MLLMs}}\\
         \cmidrule{1-12}
         Qwen-VL-Plus  &- & - & - & - & 10.7 & 38.5 & 39.1 & 39.3 & - & - & - \\
         Qwen-VL-Max   &- & - & - & 36.3 & 15.6 & - & - & - & 40.8 & 30.3 & 20.6 \\
         GPT-4V        &- & - & - & \colorbox{backred!50}{48.4} & \colorbox{backred!50}{22.8} & \colorbox{backred!50}{50.5} & \colorbox{backred!50}{53.0} & \colorbox{backred!50}{51.0} & \colorbox{backred!50}{65.5} & \colorbox{backred!50}{49.2} & \colorbox{backred!50}{38.2}\\
         \cmidrule{1-12}
         \multicolumn{12}{c}{\textit{Open-source MLLMs}}\\
         \cmidrule{1-12}
         LLaMA-Adapter V2 &7B & - & 30.6 & 23.0 & - & 25.5 & 26.3 & 24.3 & - & - & -\\
         mPLUG-Owl2 &7B & - & - & 18.8 & - & - & - & - & - & - & -\\
         UniMath &-  & 50.0 &- &- &- &- & - & - & - & - & -\\
         LLaVA-1.5 &13B & 20.3 & 21.0 & 24.0 & 11.1 & - & - & - & - & - & -\\
         ShareGPT4V &13B & - & - & - & 11.9 & - & - & - & - & - & -\\
         SPHINX-MoE &8$\times$7B & - & 33.9 & - & 14.2 & 31.2 & 31.7 & 30.5 & - & - & -\\
         G-LLaVA$^*$    &13B & 67.0 & - & - & - & 56.7 & - & - & 32.4 & 30.1 & 32.7\\
         Math-LLaVA$^*$ &13B & - & - & - & - & 57.7 & 53.0 & 56.5 & - & - & -\\
         InternLM-XC2. &7B & - & - & 30.1 & 14.5 & 63.0 & 56.6 & 62.3 & 47.0 & 33.1 & 33.0\\
         LLaVA-NeXT &110B& - & - & - & - & - & - & - & 53.7 & 36.9 & 31.5\\
        \cmidrule{1-12}
        \makecell[l]{\textbf{MAVIS-7B w/o DPO}} &7B & 66.7 & 40.3 & 39.2 & 18.6 & 63.2 & 58.3 & 63.0 &
        56.9 &
        37.1 &
         33.2\\
        \makecell[l]{\textbf{MAVIS-7B}} &7B & \colorbox{backblue!50}{68.3} & \colorbox{backblue!50}{50.0} & \colorbox{backblue!50}{42.4} & \colorbox{backblue!50}{19.2} & \colorbox{backblue!50}{64.1} & \colorbox{backblue!50}{59.2} & \colorbox{backblue!50}{63.2} &
        \colorbox{backblue!50}{57.2} &
        \colorbox{backblue!50}{37.9} &
         \colorbox{backblue!50}{34.6}\\
        \bottomrule
    \end{tabular}
    \end{adjustbox}
\end{table*}

\subsection{Quantitative Performance}
\label{s5.2}
As shown in Table~\ref{t10} for the MathVerse benchmark, MAVIS-7B achieves the best overall scores in both CoT evaluation and accuracy among open-source MLLMs with only a 7B model size, and consistently surpasses the second-best method on different problem versions. Specifically, our model surpasses the powerful InternLM-XComposer2 (7B)~\citep{dong2024internlm} by +9.3\% and ShareGPT4V (13B)~\citep{Chen2023ShareGPT4VIL} by +17.8\% CoT evaluation scores. Compared to other mathematical visual specialist, i.e., G-LLaVA (7B)~\citep{gao2023g} and the concurrent Math-LLaVA (13B)~\citep{shi2024math}, MAVIS-7B exhibits superior problem-solving capabilities with higher CoT evaluation scores of +19.5\% and +11.1\%, respectively. In addition, our model is also advantageous to the most powerful open-source MLLM series, LLaVA-NeXT~\citep{li2024llavanext-strong}, from 8B to 110B model sizes, demonstrating the math-specific proficiency of MAVIS-7B. Note that, the improvement brought by DPO (our fourth-stage training) is more apparent in CoT evaluation compared to the accuracy scores, indicating that the preference alignment learning can effectively boost the CoT reasoning capabilities.

Table~\ref{tab:mavis_comparison} showcases the performance comparison on six other mathematical benchmarks, where our model still attains remarkable performance among other MLLMs. In detail, MAVIS-7B outperforms the closed-source Qwen-VL-Max~\citep{Bai2023QwenVLAF} by +6.1\% in MMMU-Math, +3.6\% in MathVision, and around +10\% in three subsets of We-Math. Our model even exceeds GPT-4V~\citep{OpenAI2023GPT4TR} in the three mathematical categories of MathVista, indicating our problem-solving and reasoning proficiency. We also observe that, the enhancement from DPO increases from `S1' to `S3' of We-Math, which well demonstrates its benefit on math problems with more intricate reasoning steps.

\subsection{Qualitative Analysis}
\label{s5.3}

In Figure~\ref{figxxx}, we compare the mathematical problem-solving examples between MAVIS-7B and GPT-4V~\citep{openai2023gpt4v}. As presented, our model not only showcases better accuracy in understanding the geometric elements, function curves, and coordinate axes in mathematical diagrams, but also performs higher-quality step-by-step reasoning process for formula substitution and numerical calculation. This demonstrates the effectiveness of our four-stage training pipeline and automatic data engine for enhanced diagram understanding and CoT reasoning.

\begin{figure*}[t]
\centering
\includegraphics[width=0.98\textwidth]{figs/VSGPT.pdf}
   \caption{\textbf{Problem-solving Comparison of MAVIS-7B and GPT-4V.}}
\label{figxxx}
\vspace{-0.1cm}
\end{figure*}


\section{Conclusion}
In this paper, we propose MAVIS, the first mathematical visual instruction tuning paradigm for MLLMs. We first introduce two high-quality datasets by a delicate data engine, MAVIS-Caption and MAVIS-Instruct, containing large-scale diagram-language and problem-solving data. Then, we customize a three-stage training framework to progressively train the math-specific vision encoder, the diagram-language alignment, and the mathematical reasoning capabilities of MLLMs. The obtained specialist model, MAVIS-7B, achieves superior performance across different mathematical visual benchmarks, demonstrating the potential to serve as a new standard for future research.

\bibliography{iclr2025_conference}
\bibliographystyle{iclr2025_conference}

\clearpage 
\appendix

\section{Appendix}
\subsection{Related Work}
\paragraph{Visual Instruction Tuning.}
The advancement of large language models (LLMs)~\citep{brown2020language,albert24mixtral,touvron2023llama2,vicuna2023} with instruction tuning has significantly enhanced zero-shot capabilities across a range of tasks. Drawing inspiration from this, LLaMA-Adapter series~\citep{zhang2024llamaadapter,gao2023llamaadapterv2,han2023imagebind} propose a zero-initialized attention mechanism to align frozen vision encoders~\citep{Radford2021LearningTV} with LLaMA~\citep{touvron2023llama} for multi-modal learning.
LLaVA series~\citep{liu2023llava,liu2023improvedllava} employ a linear projector for vision-language alignment, establishing visual instruction tuning as a standard training approach in the multi-modal field. Flamingo~\citep{alayrac2022flamingo} and OpenFlamingo~\citep{awadalla2023openflamingo} have honed visual representation by integrating a cross-attention resampler with vision encoders. SPHINX series~\citep{gao2024sphinx,lin2023sphinx} utilize a blend of visual encoders to make the LLM cognizant of various image aspects. InternVL series~\citep{chen2024far,dong2024internlm,team2023internlm} employ a large vision encoder and QFormer~\citep{li2022blip} to incorporate high-quality visual information through a multi-stage training methodology. 
LLaVA-NexT~\citep{liu2024llavanext,li2024llavanext-strong,li2024llavanext-interleave} further introduces the `AnyRes' technique to manage images at any given resolution, and LLaVA-NexT-Interleave~\citep{li2024llavanextinterleavetacklingmultiimagevideo} extends the scope widely to interleave multi-image settings. There are also recent efforts to apply visual instruction tuning to 3D~\citep{guo2023point,xu2023pointllm} and video~\citep{li2023videochat,fu2024video} scenarios.
Despite the impressive strides made in both model capability and training efficiency by multi-modal large language models (MLLMs) through visual instruction tuning, there is currently no MLLM specifically designed for mathematical problem-solving, nor a substantial dataset available for such purposes in the open-source community. In this paper, we mitigate the issue by proposing MAVIS with high-quality mathematical visual datasets and training paradigms.

\paragraph{Mathematics in Large Models.}
Recent research has predominantly concentrated on text-only mathematical problem-solving using LLMs. MAmmoTH~\citep{yue2023mammoth,yue2024mammoth2} have compiled extensive collections of mathematical problems, training LLMs using the reasoning processes described in solutions. MetaMATH~\citep{yu2023metamath} has expanded upon this by rewriting existing problems to create a larger dataset. MathCoder~\citep{wang2024mathcoder} and ToRA~\citep{gou2023tora} introduced a tools agent approach, employing Python code and symbolic resolvers during the training phase, significantly outperforming traditional models that rely on text-only mathematical reasoning.
However, in the multi-modal field, despite the introduction of several datasets such as Geometry3K~\citep{lu2021inter}, GeoQA~\citep{Chen2021GeoQAAG}, UniGeo~\citep{Chen2022UniGeoUG}, UniMath~\citep{liang_unimath}, and GeomVerse~\citep{kazemi2023geomverse}, aiming at enhancing the performance of MLLMs in solving graphical mathematical problems, these datasets are quite limited in scale and domain. Based on these datasets, G-LLaVA~\citep{gao2023g} has developed superior capabilities for understanding graphical geometries but struggles with mathematical problems in other domains. The comprehensive benchmark MathVerse~\citep{zhang2024mathverse} has also highlighted the existing MLLMs' unsatisfactory capacity for encoding visual diagrams in diverse mathematical domains.
Therefore, there is a pressing need for the development of more robust encoders for mathematical images and the tuning of MLLMs with mathematical visual instructions, for which we propose MAVIS to address the challenges.

\subsection{Human Evaluation of MAVIS-Instruct}
To assess the dataset's coverage, validity, and quality, human verification is employed. The creation process of our MAVIS-Instruct dataset can be broadly categorized into two approaches:

\begin{itemize}
    \setlength\itemsep{0.3cm} 
    \setlength\leftskip{0cm}  
    \renewcommand\labelitemi{\raisebox{0.2ex}{\tiny$\bullet$}} 
    \item \textbf{GPT-generated:} This method leverages GPT-4 to generate new problems (including questions, rationales, and answers) based on existing problems with diagrams. While this approach produces fluent, human-like sentences, it may be influenced by the inherent capabilities and occasional instability of GPT-4V.
    \item \textbf{Data Engine:} As the main source of our mathematical visual data, this method utilizes the custom automatic data engine to generate new problems (including diagrams, questions, rationales, and answers), without relying on GPT models. It guarantees 100\% correctness due to the use of rigorous templates, though it may occasionally exhibit rigid expressions.
\end{itemize}

\begin{figure*}[t]
\vspace{-0.25cm}
\centering
\begin{minipage}[c]{0.45\textwidth}
\includegraphics[width=0.999\textwidth]{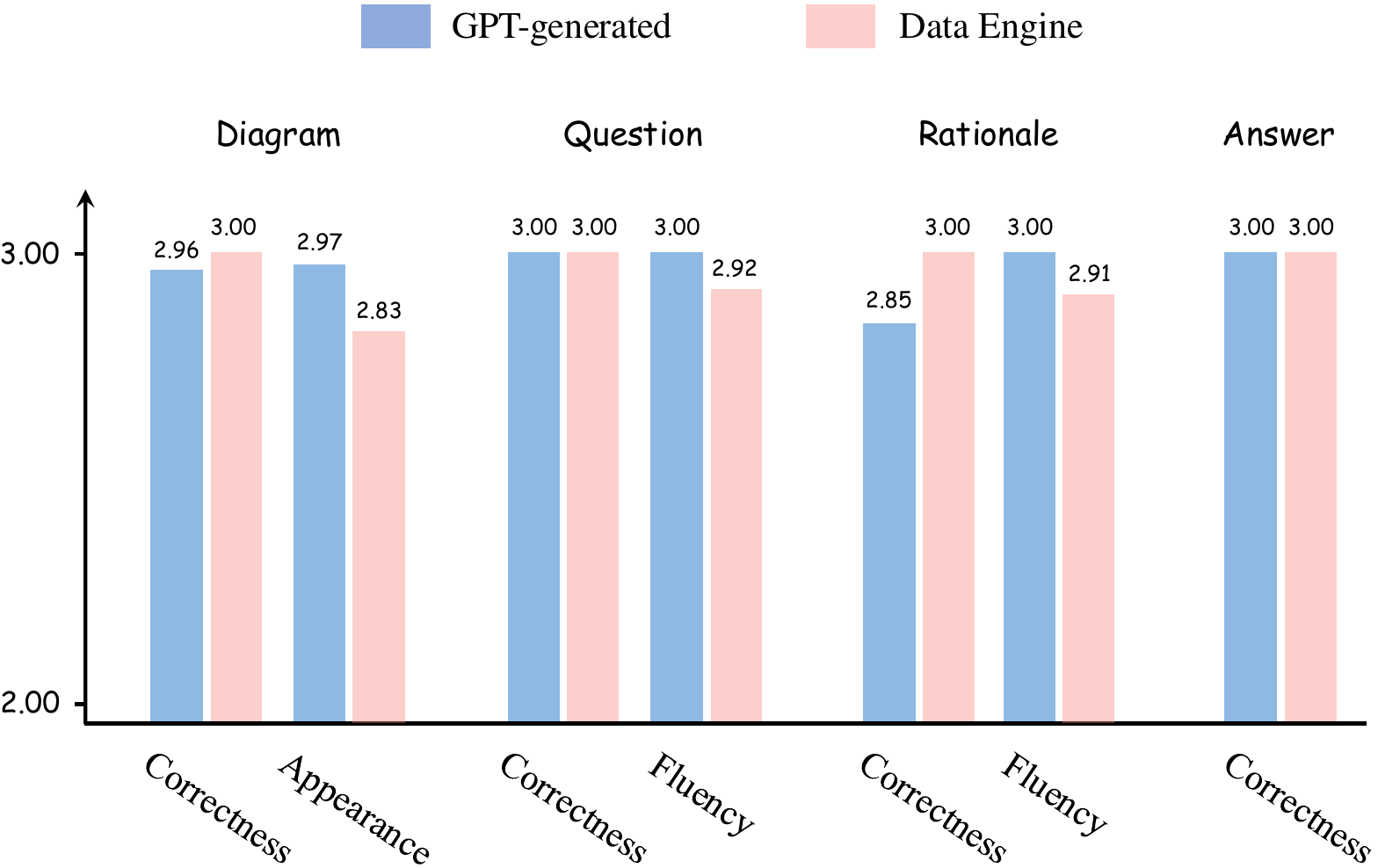}
\caption{\textbf{Human Evaluation Results} on 200 randomly sampled problems in MAVIS-Instruct, 100 GPT-generated and 100 Data Engine. We set three levels (1, 2, and 3) for each metric, and report average scores.}
\label{fig:Avg-Score}
\end{minipage}\qquad
\begin{minipage}[c]{0.45\textwidth}
\includegraphics[width=0.999\textwidth]{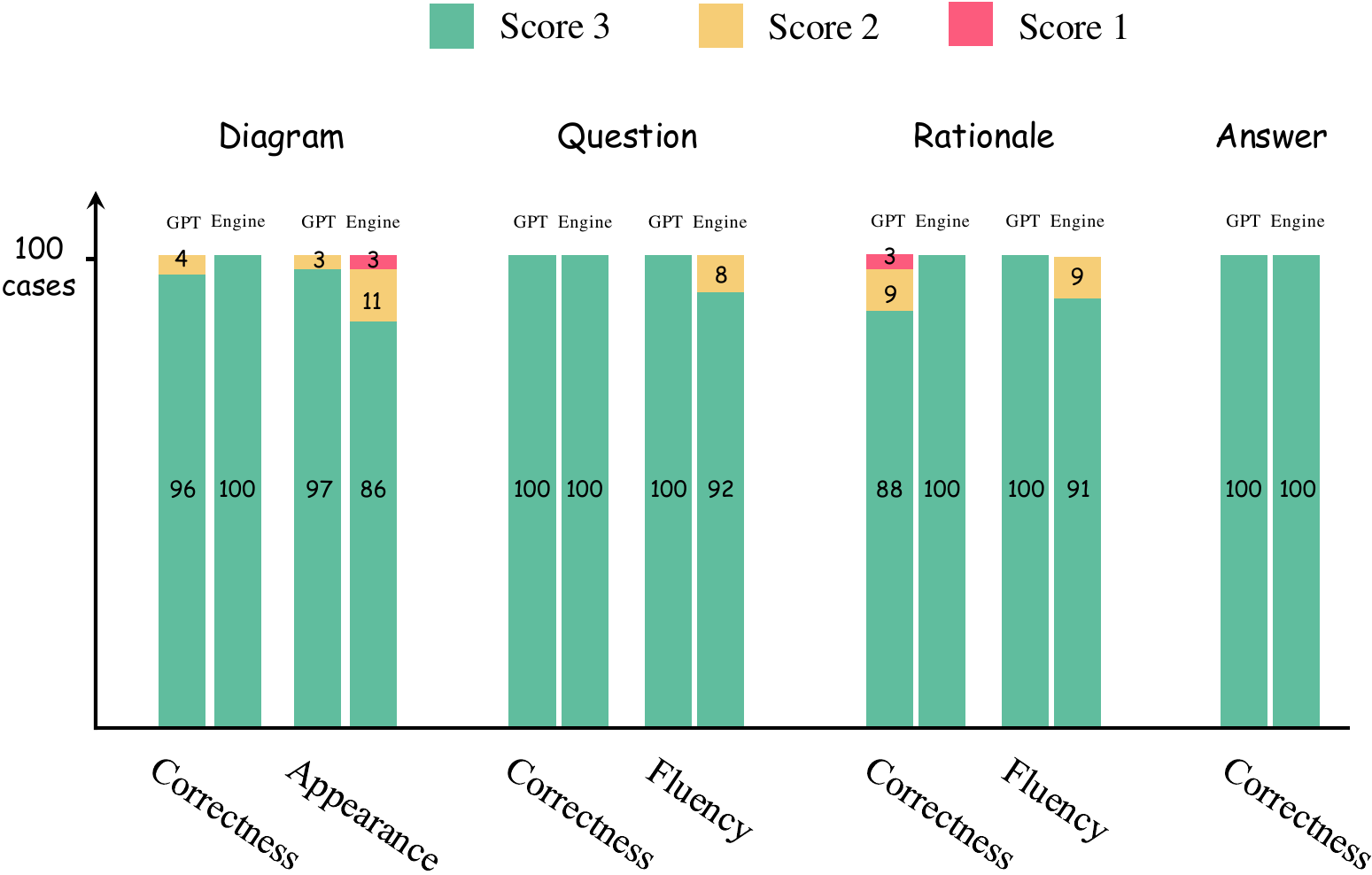}
    \caption{\textbf{Human Evaluation Statistics} on 200 randomly sampled problems in MAVIS-Instruct, 100 GPT-generated and 100 Data Engine. We count the numbers of three score levels (1, 2, and 3) for each metric.}
    \label{fig:Distribution}
\end{minipage}
\end{figure*}

\begin{figure*}[htbp]
    \centering
    \includegraphics[width=0.999\textwidth]{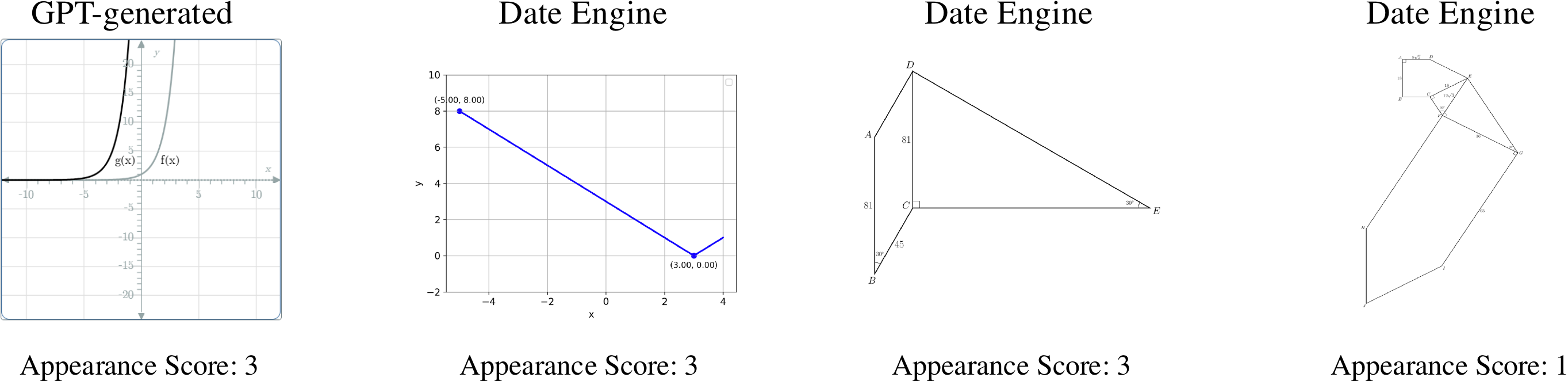}
    \caption{\textbf{Diagram Examples in MAVIS-Instruct.} The first three diagrams showcase superior correctness and appearance, while a small portion of Data Engine generated diagrams (3\%) are not aligned with human preference, e.g., the fourth diagram.}
    \label{fig:R1-Diagram}
\end{figure*}

\begin{figure*}[htbp]
\vspace{0.1cm}
\includegraphics[width=0.999\textwidth]{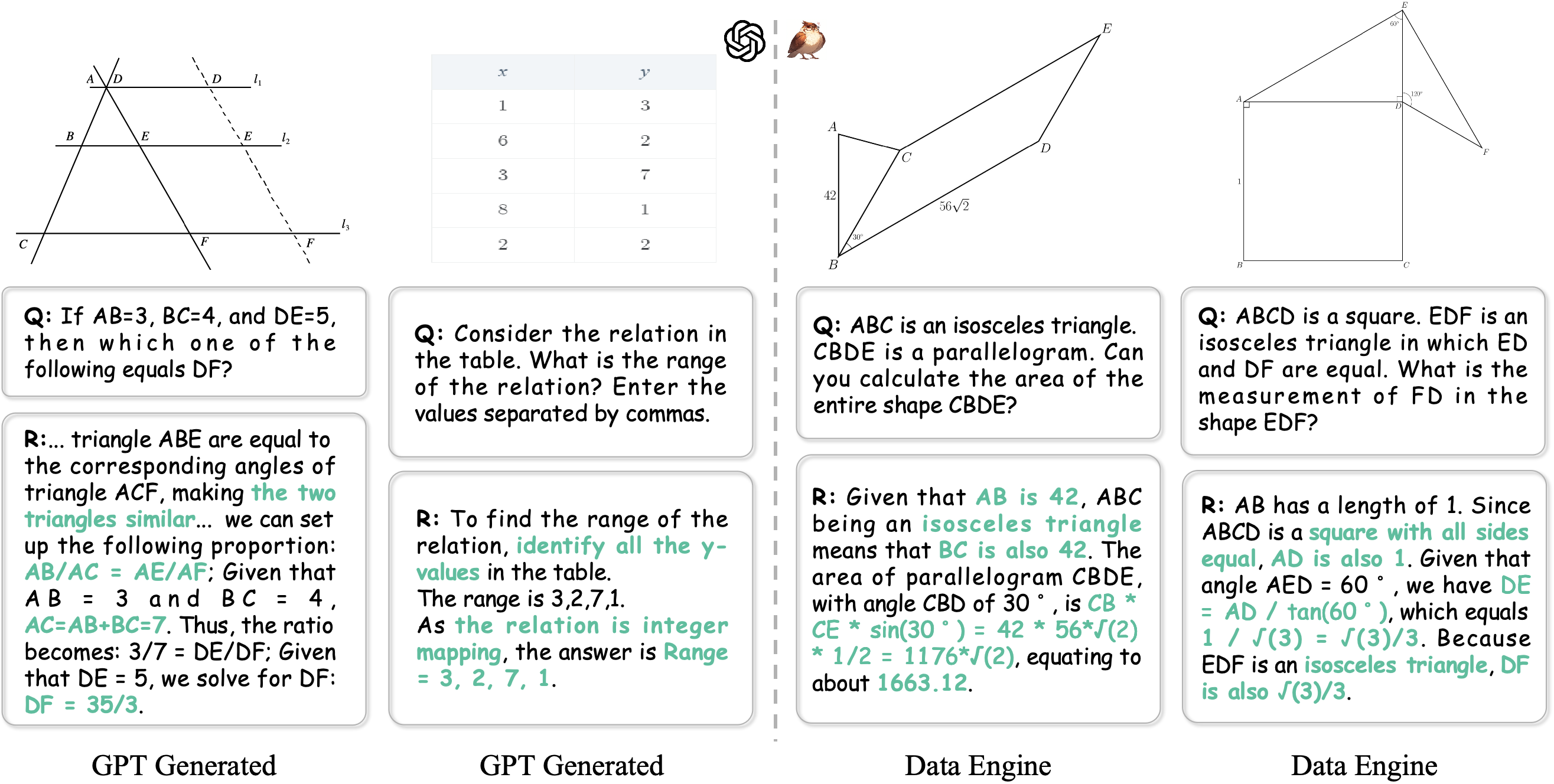}
\caption{\textbf{Accurate Rationale Examples in MAVIS-Instruct.} Most GPT-generated and Data Engine-generated rationales ensure correctness.}
\label{fig:Sample-QAR-Positive}
\end{figure*}

Specifically, we evaluate four aspects(Diagram, Question, Rationale and Answer) of each problem using seven metrics. Each metric is scored on a scale of 1 to 3, where 1 denotes \textit{poor}, 2 denotes \textit{moderate}, and 3 denotes \textit{good}. The human evaluation results are shown in Figure~\ref{fig:Avg-Score} and score statistics  are shown in Figure~\ref{fig:Distribution}. In addition, we also showcase some specific examples in Figure~\ref{fig:R1-Diagram} and Figure~\ref{fig:Sample-QAR-Positive}. We analyze each aspect as follows:

\begin{itemize}
    \setlength\itemsep{0.3cm} 
    \setlength\leftskip{0cm}  
    \renewcommand\labelitemi{\raisebox{0.2ex}{\tiny$\bullet$}} 
    \item \textbf{Diagram:} The diagrams in GPT-generated problems are directly collected from existing sources with rigorous human filtering, ensuring high quality, resulting in scores close to 3. In contrast, for rule-based problems, the diagrams are drawn accurately using Python code driven by our data engine, which guarantees correctness. However, these diagrams may lack alignment with human aesthetic preferences, as indicated by 3\% of them receiving an appearance score of 1.
    \item \textbf{Question:} Regarding the questions, both GPT-generated and rule-based problems display a high degree of accuracy in aligning with the diagram elements. This is attributed to the well-crafted prompts used with GPT-4 and the meticulous template design of the data engine. Nevertheless, rule-based questions may occasionally exhibit minor fluency issues, as they lack human refinement.
    \item \textbf{Rationale:} In terms of the rationales, most instances feature a precise and detailed chain-of-thought (CoT) reasoning process. However, in a few cases (3\% receiving an accuracy score of 1), some GPT-generated rationales contain minor reasoning or calculation errors, which are inherent to GPT-4's limitations in problem-solving. These errors usually affect only one or two steps and do not compromise the overall logic. Conversely, the rule-based rationales are highly accurate due to the carefully designed data engine, although there is still room for improvement in language fluency.
    \item \textbf{Answer:}  The answers in both methods achieve high correctness scores. For GPT-generated problems, we prompt GPT-4 to identify a known condition from the original problems as the answer. Similarly, for rule-based problems, we randomly select a known attribute from the generated diagrams to serve as the answer.
\end{itemize}

Overall, the randomly sampled instances show that our dataset exhibits good question quality and answer accuracy.

\end{document}